\pdfoutput=1

\documentclass[11pt]{article}

\usepackage[]{acl}

\usepackage{times}
\usepackage{latexsym}
\usepackage{natbib}
\usepackage{graphicx}
\usepackage{float}
\usepackage{colortbl}
\usepackage{booktabs}
\usepackage{float}
\usepackage{multirow}
\usepackage{graphicx}
\usepackage[normalem]{ulem}
\useunder{\uline}{\ul}{}

\usepackage{fancyhdr}

\usepackage{cleveref}
\crefname{section}{Sec.}{Sec.}
\crefname{Section}{Sec.}{Sec.}
\crefname{table}{Tab.}{Tab.}
\crefname{appendix_table}{Tab.}{Tab.}
\crefname{Table}{Tab.}{Tab.}
\crefname{Figure}{Fig.}{Fig.}
\crefname{figure}{Fig.}{Fig.}
\crefname{appendix}{Appendix}{Appendix}
\crefname{chapter}{Chapter}{Chapter}

\makeatletter
\renewcommand{\paragraph}{%
  \@startsection{paragraph}{4}%
  {\z@}{0.75ex \@plus 1ex \@minus .2ex}{-1em}%
  {\normalfont\normalsize\bfseries}%
}
\makeatother

\usepackage[T1]{fontenc}

\usepackage[utf8]{inputenc}

\usepackage{microtype}

\newcommand\blfootnote[1]{%
  \begingroup
  \renewcommand\thefootnote{}\footnote{#1}%
  \addtocounter{footnote}{-1}%
  \endgroup
}

\title{Looking for a Handsome Carpenter! \protect\\ Debiasing GPT-3 Job Advertisements}

\fancypagestyle{firstpage}
{
    \fancyhead[L]{}
    \fancyhead[CO]{This paper is accepted for the 4th Workshop on Gender Bias in Natural Language Processing at NAACL 2022.}
    \fancyhead[R]{}
}

\author{
{\small Conrad Borchers$^{\dagger\ddagger*}$, Dalia Sara Gala$^{\dagger*}$, 
Benjamin Gilburt$^{\dagger*}$,}\\
{\small \textbf{ Eduard Oravkin$^{\dagger}$, Wilfried Bounsi$^{\dagger}$, Yuki M. Asano$^{\dagger}$, Hannah Rose Kirk$^{\dagger}$}} \\
  {\small  $^\dagger$Oxford Artificial Intelligence Society, University of Oxford} \\
  {\small $^\ddagger$conrad.borchers@oii.ox.ac.uk} \\
}

\begin{document}
\maketitle
\thispagestyle{firstpage}
\begin{abstract}
The growing capability and availability of generative language models has enabled a wide range of new downstream tasks. Academic research has identified, quantified and mitigated biases present in language models but is rarely tailored to downstream tasks where wider impact on individuals and society can be felt. In this work, we leverage one popular generative language model, GPT-3, with the goal of writing unbiased and realistic job advertisements. We first assess the bias and realism of zero-shot generated advertisements and compare them to real-world advertisements. We then evaluate prompt-engineering and fine-tuning as debiasing methods. We find that prompt-engineering with diversity-encouraging prompts gives no significant improvement to bias, nor realism. Conversely, fine-tuning, especially on unbiased real advertisements, can improve realism and reduce bias. 

\end{abstract}
\blfootnote{$^*$ Equal contribution.}

\begin{figure}[t]
    \centering
    \includegraphics[width=0.95\columnwidth]{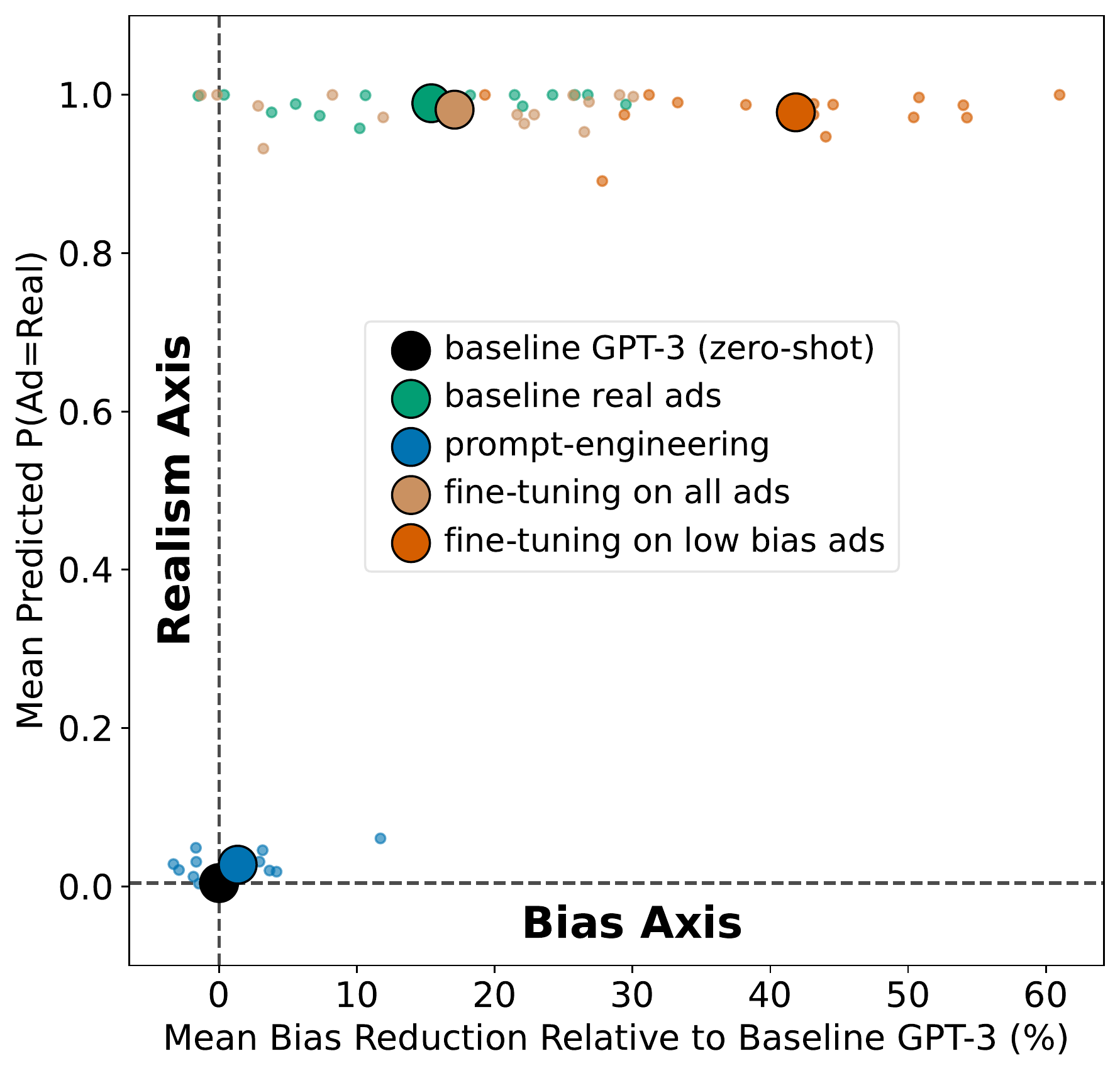}
        \vspace{-1em}
    \caption{\textbf{GPT-3 can write realistic and less biased job advertisements}. While the naïve GPT-3 zero-shot baseline is both highly biased and easily identified as synthetic, prompt-engineering and more importantly fine-tuning on real and less-biased data can substantially increase realism and decrease bias.}
    \label{fig:splash}
\end{figure}

\section{Introduction}
Generative language models are getting bigger: from ELMo's release in 2018 with 94M parameters \cite{Joshi2018} to Megatron-Turing NLG in 2022 with 530Bn \cite{Smith2022}, there has been approximately a tenfold annual increase in parameters. The growing capabilities of these models have supported their adoption in many downstream tasks, from text summarisation \cite{Li2020} and weather reporting \cite{Gatt2018} to writing code \cite{Chen}. However, there are various associated risks, such as privacy erosion, copyright infringement, environmental harms and negative stereotyping of social groups \cite{Margoni2019, Feyisetan2020, Bender2021, bommasani2021opportunities, Weidinger2021}. We focus on the latter of these risks, specifically the problem of gender bias with respect to occupation. 

The reward and risk of using generative models in tasks related to job search are debated. While some argue for the value of text generation and summarisation technologies to promote inclusive hiring \cite{10.3115/974557.974597}, others suggest model biases towards occupational associations pose a risk of their use. Specifically, research has uncovered gender bias in large-scale language models by examining the strength of statistical association between a given gender and a set of jobs using prompts such as ``the woman works as a [token]'' \cite{Sheng2019, Kirk2021}. These associations lead to representational harms \cite{Blodgett2020}, by perpetuating the notion of gendered roles in the labour force and entrenching stereotypes such as women possessing more caregiving qualities. However, it is unclear how these model biases translate directly to language generation in applied downstream tasks; that is, how they may give rise to allocational harms. One example of such a task is the generation of job advertisements (ads) which exemplifies the risk of allocational harms because candidates from a given group may be discouraged to apply as a result of biased language. Prior research has demonstrated gendered wording in job ads can act as an institutional-level mechanism to entrench traditional gender divisions \cite{Gaucher2011}.\footnote{In our experiments, GPT-3 began one ad with ``Handsome carpenter with an eye for detail needed'', where \textit{handsome} is defined as ``physically attractive (esp. of a man)'' \cite{_handsome}.}  

Gender bias in natural language processing (NLP) has been more widely-discussed \cite{sun2019mitigating, Blodgett2020, lu2020gender}, with some specific work documenting bias of generative language models \cite{solaiman2019release, brown2020language, Kirk2021}. Early debiasing attempts in NLP focused on word embeddings \cite{Bolukbasi2016, Kurita2019}, though the efficacy of these methods has been challenged \cite{Gonen2019}. Some recent research seeks to align generative language models with societally-desirable values \cite{solaiman2021process}, reduce various dimensions of group-directed bias \cite{ liu2021mitigating, smith2021hi} and decrease risk of toxicity \cite{ouyang2022training}. There is less research on how gender bias in generative models affects applied tasks, and to our knowledge, no prior work on bias in generated job ads. Furthermore, there is a lack of research advising on how industry practitioners can effectively and cheaply debias outputs whilst retaining quality, accuracy and realism. 

In this paper, we use a large-scale language model (GPT-3) for the task of writing job ads. Our goal is to generate job ads that are (1) \textit{unbiased}, i.e., do not encourage or discourage application from one gender; and (2) \textit{realistic}, i.e., of a quality comparable to human-generated ads. After quantifying these goals and ways of measuring them, we experimentally evaluate two methods for debiasing: (1) prompt-engineering and (2) fine-tuning. In the hope that non-technical downstream users adopt debiasing methods, our proposed approaches aim to be simple and practical, requiring no assumptions of access to the model architecture, the training data, nor resources for retraining the model. We find that, compared to a zero-shot baseline, prompt-engineering improves neither bias, nor realism (\cref{fig:splash}). This is an important discovery because prompt-engineering is one of the easiest ways that a practitioner could try to mitigate bias, for example by simply modifying ``Write a job ad for a carpenter'' to become ``Write a job ad for a carpenter \textit{for
a firm focused on diversity in
hiring}''. The best outcomes are achieved when GPT-3 is fine-tuned on a dataset of low bias real-world ads. This paper provides the following contributions:
\begin{itemize}
\setlength\itemsep{-0.4em}
\item A method for using GPT-3 in an applied scenario of generating job ads which, to our knowledge, has not been researched before.
\item A composite score-based quantification of text-level gender bias in job ads.
\item A comparative study of real-world job ads and those created by GPT-3 in a zero-shot, prompt-engineered and fine-tuned setting, evaluated w.r.t. bias \textit{and} realism.
\end{itemize}

\section{Bias Statement}

In this paper, we focus on measuring and mitigating gender-biased language in machine-generated job ads, a use case of large-scale language models which risks representational and allocational harms \cite{Blodgett2020}. Representational harms come from the conditioning of a job's suitability to a given individual based on their gender. When jobs are valued unequally (either by financial, social or intellectual status), this, in turn, can reinforce gendered power hierarchies and negative societal divisions. Gender-biased language may result in an unequal distribution of job applications if it dissuades gender-diverse candidates from applying \cite{Gaucher2011}. Thus, allocational harms are relevant where labour market opportunities, financial remuneration or job stability are preferentially granted based on gender. We know from prior NLP research that GPT models reflect occupational stereotypes in society \cite{Kirk2021}, confirming the risk of representational harm, but not how this translates into allocational harms in applied settings. To measure bias, our experiment employs lists of gender-coded words. These lists are potentially in themselves biased, having been defined by a research group under a particular cultural bias or as the result of biased data. To mitigate this concern, we use multiple measures to cover the blind spots or specific biases present in any single list. However, our proposed metric may better capture the most obvious, text-level aspects of gender-biased language and will be less effective to find covert, but equally as damaging, forms of gender bias in job ads, or job search more broadly.

\section{Methods}
\label{sec:methods_intro}
We define our task as generating job ads, i.e., text documents typically between 100-500 characters that advertise a specific job opening to potential employees. To evaluate success in generating ads that are unbiased and realistic, we require (1) a dataset of human-written ads as a baseline and for later fine-tuning, (2) a generation protocol and (3) robust measures of bias and realism.
\subsection{Data Collection and Generation}
\label{sec:real_ads}

\paragraph{Job Selection} Collecting and generating job ads for all possible jobs is prohibitively timely and costly. Hence, we restrict our experiments to a sample of 15 jobs selected via three criteria: (1) \textit{prevalence}, jobs with a sufficiently large labour force in the UK ($N\ge40{,}000$), (2) \textit{relevance}, jobs which have a sufficiently large number of real-world job ads on a popular online forum ($N\ge1{,}000)$ and (3) \textit{bias}, jobs which represent the most male-biased, female-biased and neutral parts of GPT-3's prior distribution in how frequently certain jobs are associated with a given gender. To apply these criteria, we first filter jobs in the UK economy by prevalence and relevance \cite{ONS2018}. Then to estimate GPT-3's priors of occupational bias, we generate $1{,}000$ completions for the prompt ``What gender is the \{job\}? The \{job\} is a [token]'', where a completion could be: ``What gender is the plumber? The plumber is a [woman]''. Using the ratio of male-to-female tokens in these $1{,}000$ completions, we select the top five male-biased, female-biased and neutral jobs (see \cref{sec:job_selection_detail} for further detail on job selection and pre-processing).\footnote{\textbf{Male-biased jobs:} plumber, engineer, carpenter, electrician, software developer; \textbf{Female-biased jobs:} nurse, housekeeper, occupational therapist, secretary, social worker; \textbf{Neutral jobs}: artist, tester, administrator, project manager, writer.}
\paragraph{Collecting Real-World Ads} 
To generate a dataset of human-written ads, we collect live job ads matching the 15 selected job titles from a popular UK job site in January 2022. 
After de-duplication, our sample includes 85 ads per job.

\paragraph{Generating Job Ads} We use the OpenAI Davinci GPT-3 model which has been adapted for natural language requests. We use default parameters values and 500 maximum tokens per completion (see \cref{sec:gpt_hyperparams} for hyperparameter details). Keeping default parameters better reflects when non-technical users apply large-scale generative models ``out-of-the-box'' \cite{Kirk2021}. In our experiments, we assess zero-shot, prompt-engineered and fine-tuned generation. We use templated prompts e.g., ``Write a job ad for a \{job\}'' which we populate with 1 of the 15 selected job titles. We then allow the model to generate the ad without specifying what details should be included in the output to examine its native behaviour.

\subsection{Experimental Conditions} 
We evaluate debiasing methods which could feasibly be implemented by practitioners in an applied setting. Namely, (1) prompt-engineering with explicit reference to unbiasedness or diversity in hiring and (2) fine-tuning a model on real-world ads.\footnote{We also tried combining prompt-engineering and fine-tuning but it worsened realism and bias.} For each condition, our bias and realism metrics are calculated at the document level, then averaged over female-biased, male-biased and neutral jobs. We define the bias metrics in \cref{sec:measuring_bias} and realism metrics in \cref{sec:measuring_realism}. For each condition, we create a dataset of ads and remove any duplicates (see \cref{tab:summary_table} for experimental datasets and the number of ads they contain).

\paragraph{Baselines}
\label{sec:baselines}
We define two baseline datasets of job ads. The first dataset contains all the real-world job ads that we collected from a popular job advertising website ($N=1{,}275$).\footnote{We assume that ads in this dataset are human-authored.} The second dataset contains ads generated by GPT-3 using zero-shot generation with nine syntactically-varied but neutral prompts such as ``Write a job ad for a \{job\}'' (see \cref{sec:prompt_detail} for prompts). For each prompt, we generate 40 completions per job title ($N=5{,}400$).

\paragraph{Prompt-Engineering}
\label{sec:prompt_engineering_methods}
In contrast to the neutral prompts in the zero-shot generative baseline condition, we define a set of prompts that explicitly attempt to force the generation of an unbiased ad. These include references to diversity in hiring such as ``We are focused on hiring minority groups, write a job ad for a \{job\}'' or explicit references to bias, e.g., ``Write a job ad for a \{job\} which appeals equally to men and women'' or ``Compose an unbiased job ad for a \{job\}'' (see \cref{sec:prompt_detail}). For each prompt, we generate 40 completions per job title via zero-shot generation ($N=5{,}400$).

\paragraph{Fine-Tuning}
\label{sec:fine_tuning_methods}
We construct three fine-tuning datasets from the real-world job ads: (1) the \textbf{all jobs dataset} has all real-world job ads for the 15 selected jobs ($N=1{,}163$). (2) the \textbf{low bias dataset} includes the 10\% least biased real ads for each job title ($N=127$), as measured by our bias metric. (3) the \textbf{high bias dataset} conversely uses the 10\% most biased real ads ($N=125$). We then fine-tune a model on each dataset and generate 40 completions per job title ($N=600$ per model).

\subsection{Measuring Bias}
\label{sec:measuring_bias}
Gender bias in language is complex and no single measure can capture all presentations of societal harms \cite{Blodgett2020}.
Several methodologies to measure and mitigate bias cannot be applied in our setting given the lack of public access to GPT-3's model architecture or training dataset, and the enormous resources needed to retrain the model from scratch. In particular, this includes training data augmentation \cite{sen2021does}, adjusting model behaviour via adversarial learning \cite{zhang2018mitigating, berg2022prompt}, and amending model embeddings \cite{dev2019attenuating}. Our analysis instead focuses on the text-level bias of model-generated outputs which we measure via a composite score based on the prevalence of certain gender-laden terms, and debiasing methods which require no access to the model architecture, nor original training data.

We define text-level bias as the frequency of certain words which are recognised as favouring one gender over another. The problem is then in defining this list of words. To avoid overfitting to one axis of gender bias, we construct a composite score based on pre-existing lists which have in turn been defined through experiments and empirical assessments \cite{Schmader2007, Gaucher2011, Sap2017, Stanczak2021}. The presence of words which are more likely to be associated with one gender does not directly result in biased outcomes. Bias may be more accurately measured as the relative gender distribution of applicants who apply to a given ad. In this work, we focus on gendered word lists as one overt presentation of gender bias but encourage further research to empirically measure allocational harm, so long as any experiments consider the ethical issues of posting ``fake'' ads online. 

\paragraph{Gendered Word Lists}
We develop our bias measure using dimensionality-reduction over six existing lists of gender-laden words: \textbf{(1, 2) Gender-Coded Word Prevalence:} \citet{Gaucher2011} define masculine-and-feminine-themed words from an experiment on job ads that discouraged female applicants. \textbf{(3) Superlative Prevalence:} \citet{Schmader2007} assess the relative frequency of positive and negative superlatives used to describe male versus female job candidates in recommendation letters. We use an established set of superlative words \cite{Veale2016}. \textbf{(4) Gender-Laden Scoring:} \citet{Sap2017} analyse 32 properties related to a set of norms to score $2{,}311$ words based on their ``gender-ladenness''. \textbf{(5) Connotation Frames:} \citet{Sap2017} define linguistic markers of power and agency associated with female versus male characters in modern films. \textbf{(6) NRC VAD Lexicon:} \citet{mohammad2018obtaining} presents a lexicon of words coded by valence, arousal, and dominance whose interpretation may interact with gender.\footnote{For a fair comparison, we implement unweighted word count measures for each list. See \cref{sec:bias_detail} for further detail and mathematical definitions.}

\paragraph{Dimensionality Reduction} We employ principal component analysis (PCA) on the six bias measures on real-world job ads to collapse them into interpretable components. We then replicate the PCA on synthetic job ads (zero-shot) and project all data points onto the first two principal components of real job ads and vice versa.

\begin{figure*}[htp]%
    \centering
    \includegraphics[scale=0.45]{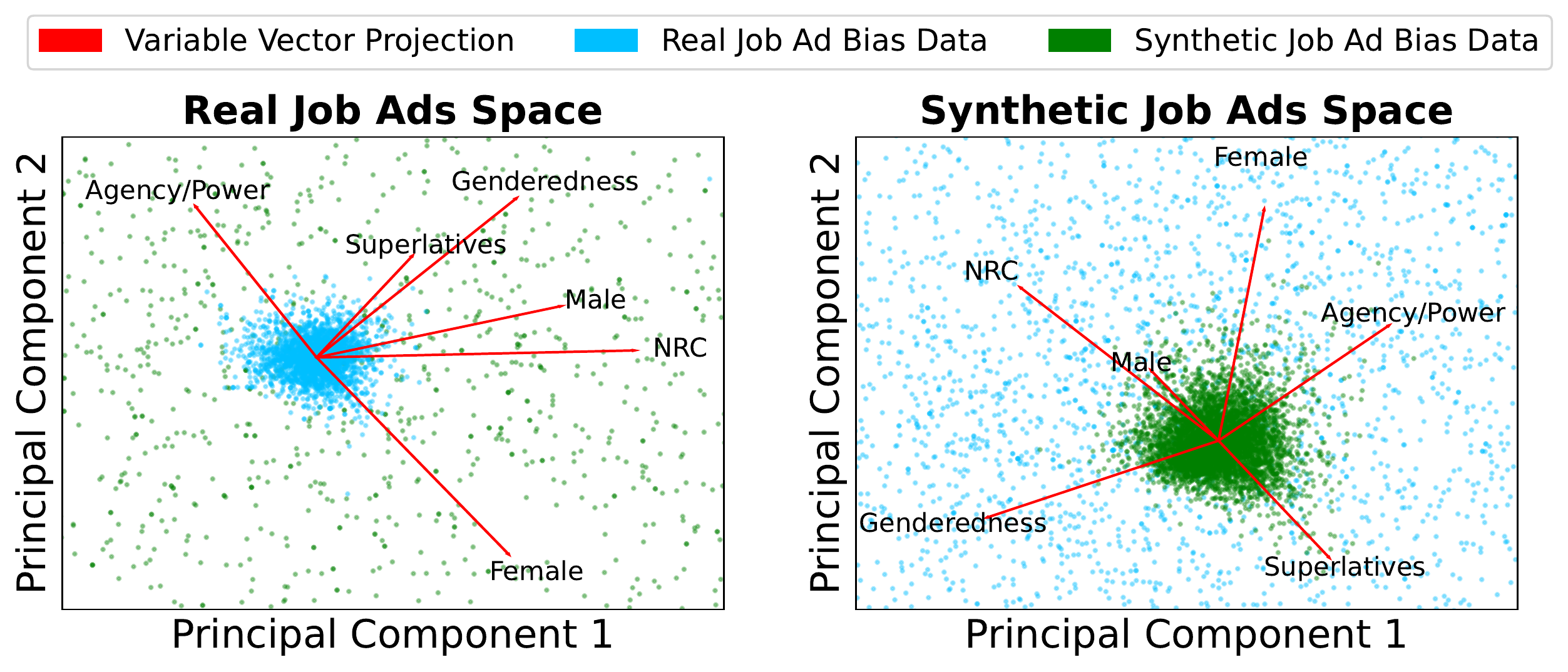}
    \caption{\textbf{Dimensionality reduction results.} Reciprocal projections of word count bias measures onto the first two principal components for real (left) and synthetic job ads created via the baseline zero-shot GPT-3 model with neutral prompts (right).\label{fig:pca_plot}} 
\end{figure*}

\subsection{Measuring Realism} 
\label{sec:measuring_realism}
We define realism as the inability to distinguish between human- and machine-generated ads. Human annotators are best placed to assess realism \cite[e.g. see][]{brown2020language} but employing and paying them to assess over $10{,}000$ ads was not feasible. Therefore, we train a discriminator model tasked with the binary prediction of whether a given input text was generated by a human or GPT-3 and validate a sample of ads using human annotators. Real ads were longer ($M=2{,}846$ characters, $SD=2{,}038$) than generated ones ($M=514, SD=210$) so we truncate texts to 500 characters. For prediction, we use a Multinominal Naive-Bayes (MNB) model, which we train, validate and test using an 80:10:10 split taken from the real and generated ads (described in \cref{sec:baselines}).\footnote{We also experimented with a BERT model \cite{devlin2018bert} but found little gain in performance to offset the greater computational and memory resources.} For our realism metric, we then use this model's predicted probability that an ad is real. To assess the robustness of this metric, we randomly sample 10 ads from each job category (female-biased, male-biased and neutral) for each experimental condition ($N=150$). We then ask three independent annotators to label the ad for whether it was human- or machine-generated and take the majority vote.\footnote{In 86\% of cases, all three annotators agreed and the Fleiss-Kappa score for inter-annotator agreement was 0.81, indicating ``very good'' agreement \cite{fleiss1971measuring}.} The accuracy of the majority label compared against the ground truth ad origin (real-world or GPT-3 generated) proxies ad quality and realism.

\section{Results}
\subsection{Dimensionality Reduction}
Employing PCA on our bias measures for real job ads results in two components, explaining 28\% and 18\% of the data variance. As shown in \cref{fig:pca_plot}, several measures representing male bias have similar vector projections. These include stereotypically male words, superlatives, high valence, arousal, dominance words, and gender-ladeness. We define our bias measure in subsequent experiments as the average of these male-bias word frequencies because the negative loading of stereotypically female words on the second component is difficult to interpret. Notably, the PCA model trained on real job ads does not replicate synthetic job ads, as demonstrated by the uncorrelated data point projection of real job ads on the right panel in \cref{fig:pca_plot}.

\subsection{Debiasing Experiments}

\begin{table}
\centering
\footnotesize
\caption{\textbf{Debiasing experiment results compared to baselines}. Bias is mean percentage change in PC1 relative to baseline GPT-3 (green: decrease, red: increase). Realism is the mean predicted probability of $ad=real$ from MNB model (Machine) and the mean predicted label of $ad=real$ from majority vote with three annotators over a sample of $30$ ads from each experiment (Human; blue: less realistic, yellow: more realistic).}
    \vspace{-0.5em}
\label{tab:summary_table}
\setlength{\tabcolsep}{2pt}
\setlength{\extrarowheight}{0pt}
\addtolength{\extrarowheight}{\aboverulesep}
\addtolength{\extrarowheight}{\belowrulesep}
\setlength{\aboverulesep}{0pt}
\setlength{\belowrulesep}{0pt}
\begin{tabular}{lcccc} 
\toprule
                        & \multicolumn{1}{l}{\textbf{Bias}}         & \multicolumn{2}{c}{\textbf{Realism }}                                               & \multicolumn{1}{l}{}  \\
\textbf{Experiment}     & \textbf{PC1}                              & \textbf{Machine}                         & \textbf{Human}                           & \textbf{$N$}          \\ 
\hline
Baseline (GPT-3)        & 0.0                                       & {\cellcolor[rgb]{0.643,0.761,0.957}}0.00 & {\cellcolor[rgb]{0.643,0.761,0.957}}0.00 & 5400                  \\ 
\hline
Baseline (Real Ads)     & {\cellcolor[rgb]{0.757,0.898,0.827}}-15.4 & {\cellcolor[rgb]{1,0.898,0.6}}0.99       & {\cellcolor[rgb]{1,0.898,0.6}}1.00       & 1275                  \\
Prompt-Engineering      & {\cellcolor[rgb]{0.976,0.988,0.984}}-1.3  & {\cellcolor[rgb]{0.651,0.765,0.957}}0.03 & {\cellcolor[rgb]{0.643,0.761,0.957}}0.03 & 5397                  \\
Fine-Tuning (All)       & {\cellcolor[rgb]{0.729,0.89,0.812}}-17.1  & {\cellcolor[rgb]{1,0.937,0.745}}0.98     & 0.95                                     & 600                   \\

Fine-Tuning (Low Bias)  & {\cellcolor[rgb]{0.341,0.733,0.541}}-41.8 & {\cellcolor[rgb]{1,0.953,0.808}}0.98     & {\cellcolor[rgb]{1,0.898,0.6}}1.00       & 600                   \\
Fine-Tuning (High Bias) & {\cellcolor[rgb]{0.902,0.486,0.451}}+9.0  & {\cellcolor[rgb]{0.992,0.996,0.996}}0.96 & 0.90                                     & 596                   \\
\bottomrule
\end{tabular}
\end{table}

\begin{figure*}[htp]
    \centering
    \includegraphics[width = \textwidth]{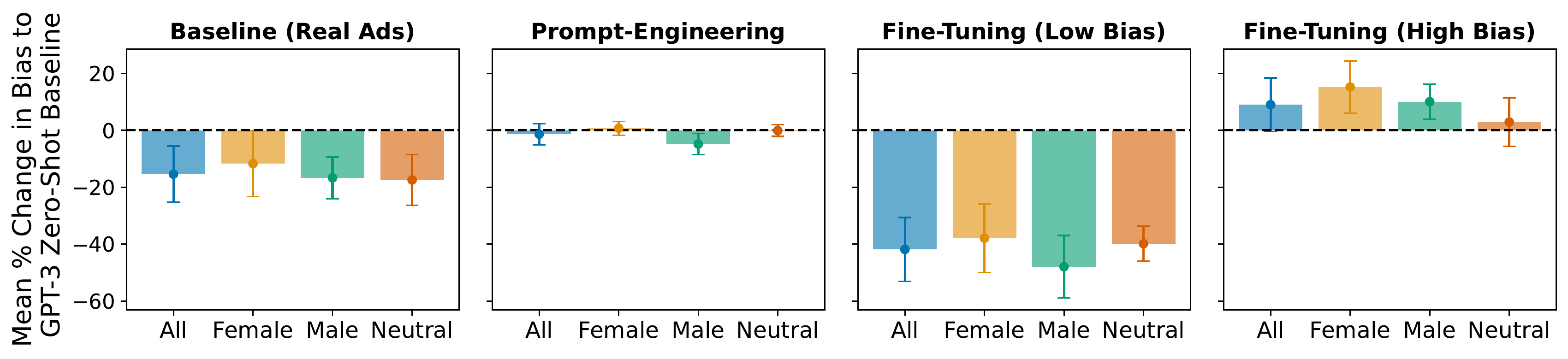}
        \vspace{-2em}
    \caption{\textbf{Bias reduction comparison of various methods}, stratified by job category: all jobs, female-biased, male-biased and neutral. While real job ads are less biased than the GPT-3 zero-shot condition, fine-tuning with low bias ads can reduce bias even further. As expected, fine-tuning on high-biased job ads increases bias. The error bars represent the standard deviation of job means within categories.}
    \label{fig:barplots}
\end{figure*}

\paragraph{Prompt-Engineering}
\label{sec:prompt_engineering_results}
Prompt-engineering does not effectively lower bias nor increase realism in generated ads (see \cref{tab:summary_table} and \cref{fig:splash}), apart from a small but significant reduction in bias for male-dominated jobs (\cref{fig:barplots}). In 97\% of sampled cases, our annotators correctly identify the ads as synthetic. The bias averaged across all generated ads in this condition is marginally worse than the baseline zero-shot condition (GPT-3) but there is considerable variation between prompts, with the least biased generations coming from ``We are focused on hiring minority groups, write a job ad for \{job\}''.\footnote{See \cref{sec:prompt_detail} for full results per prompt.}

\paragraph{Fine-Tuning}
We find that fine-tuning on real ads increases the length of generated ads, with an average of 260 words compared to 82 words in the zero-shot baseline. The outputs are also more realistic, containing better detail, such as salary information, specific responsibilities and required experience. Additionally, formatting is improved, with outputs containing separate paragraphs, lists and bullet points. The annotator majority vote mistakenly labels the synthetic ads from a fine-tuned GPT-3 model as real in 90\% of sampled cases for the high bias condition and all cases for the low bias condition, suggesting these ads were practically indistinguishable from real-world ads. Specifically, fine-tuning on low bias ads results in a significant bias reduction across all job types (\cref{fig:barplots}). This reduction in bias even outperforms the average bias of real job ads, yet retains realism (\cref{fig:splash}).

\section{Discussion \label{sec:discussion}}
Our main contributions are (1) an application of GPT-3 to a specific applied scenario with a risk for allocational harms, (2) a composite text-level measure of gender bias in this scenario relying on general and job market specific word lists and (3) preliminary findings regarding the relative success of prompt-engineering versus fine-tuning for debiasing job ads. Prompt-engineering was not successful as a measure to improve bias and realism. Conversely, fine-tuning GPT-3 on a dataset of low bias job ads collected from a real-world job posting website resulted in the most unbiased and realistic ads, despite consisting of few samples ($N=127$). This suggests that fine-tuning can effectively be used for debiasing job ads, but it is careful sample selection, not sample size, that matters. Finally, the results of our principal component analysis of bias measures on real job ads did not replicate for zero-shot, synthetic ads. Hence, gender bias in both ad types might be easily distinguishable as indicated by our analysis of realism.

\subsection{Limitations and Future Work}
\paragraph{Measurements} Our measures of bias and realism are relatively simplistic. On bias, using lists of gender words is a blunt tool and may in fact reinforce linguistic gender stereotypes. Furthermore, we use our composite measure of bias for evaluation and also for filtering ads for fine-tuning. Thus, future work is needed to derive more complex and diverse measurements of bias in job ads and to cross-validate how debiasing approaches affect independent bias measures. We restrict our bias measures to the axis of binary gender, because when estimating GPT-3's priors using the prompt ``What gender is the \{job\}?'', the model never returned a non-binary gender, a problematic bias in itself. Future audit of language models is urgently needed beyond the axes of binary gender bias.  

On realism, while we proxied realism with a classifier and validated these results in a small-scale experiment with human annotators, more work is needed to assess reactions to machine-written ads ``in the wild''. Furthermore, while fine-tuning and prompt-engineering increased realism in the aggregate, some job ads were still nonsensical or simply parroted the prompt text, e.g., ``The job ad should not have any biases in it.''. We briefly assess some outputs qualitatively in \cref{sec:bizarre_detail} but make our bias measure generation process publicly available to encourage more human-directed assessments of bias and realism.\footnote{\url{https://github.com/oxai/gpt3-jobadvert-bias}} It remains to be seen whether realism (as measured by similarity to human-authored ads) is a necessary characteristic for success (as measured by the number of applications). Prior research identifies fluency and a clear presentation of relevant skills and experience as relevant to the creation of a ``good'' job ad \cite{liu-etal-2020-hiring}, but it is not clear whether an ad must appear human-written to achieve this. Our assumption for this project is that human-written job ads follow styles, conventions and a level of detail that effectively encourage prospective employees to apply, but further research is required to understand whether ads clearly identified as machine-written can be equally or more effective in this regard. 

\paragraph{Domain} Our chosen domain of generative job ads is unlikely to be a widely used application of GPT-3 in the near future. While the computational cost of generating a single job ad is significantly lower than a human writing an ad, the human cost of reviewing generated ads and adapting them to company-specific requirements likely diminishes the cost savings. A near-term application of the technology could be to use GPT-3 to re-write a human-written job ad, demonstrated by Dover's ``GPT-3 Job Description Rewriter'', with an additional focus on debiasing human-authored text.\footnote{\url{https://producthunt.com/posts/gpt-3-job-description-rewriter-by-dover}} Our findings demonstrate that generative models must be carefully applied when creating texts for a downstream, real-world setting in hiring and recruitment, especially when used zero-shot with no debiasing techniques. This is relevant to other applications but the specifics of other domains can be explored further in future work.

\paragraph{Impact on Job Applications} While our goal was to generate gender-neutral job ads, it remains possible that neutrality may still dissuade a particular group from applying \cite{Gaucher2011}. Our work cannot comment experimentally on whether less-biased ads at the text-level result in a greater diversity of applicants. Further social science and experimental research is thus necessary to understand the effects that language in job ads has on applications from various protected groups. 

\paragraph{Generalisability} While we have established methods for measuring and mitigating binary gender bias, we have not achieved the same for non-binary genders nor for any other protected characteristics defined in the Equality Act 2010 \cite{fell2017against}. Practitioners tackling more varied presentations of identity-directed bias may be less able to find pre-existing lists of biased words to define bias measurements. 

\section{Conclusion}
To conclude, fine-tuning on a pre-selected sample of low bias job ads from real job market sites may be an effective and resource-friendly way of reducing gender bias in GPT-3 generated job ads while remaining realistic to human-authored text. Meeting both of these goals is required for the successful \textit{and} safe adoption of generative language models for downstream tasks in domains which risk allocational harms, such as hiring and job search.

\section*{Acknowledgements}
We thank our anonymous reviewers for their helpful feedback and William Lee for his contributions to the ideation and design of this study. Hannah Rose Kirk is supported by the UK Economic and Social Research Council grant ES/P000649/1. 

{\footnotesize \bibliographystyle{acl_natbib} \bibliography{main}}

\begin{thebibliography}{42}
\expandafter\ifx\csname natexlab\endcsname\relax\def\natexlab#1{#1}\fi

\bibitem[{Bender et~al.(2021)Bender, Gebru, McMillan-Major, and
  Shmitchell}]{Bender2021}
Emily~M. Bender, Timnit Gebru, Angelina McMillan-Major, and Shmargaret
  Shmitchell. 2021.
\newblock \href {https://doi.org/10.1145/3442188.3445922} {On the dangers of
  stochastic parrots: Can language models be too
  big?\raisebox{-5pt}{\includegraphics[scale=0.075]{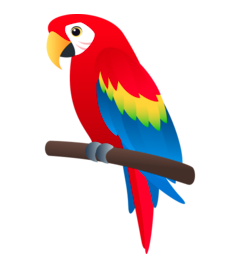}}}.
\newblock In \emph{Conference on Fairness, Accountability, and Transparency
  (FAccT ’21)}. ACM, New York, NY, USA.

\bibitem[{Berg et~al.(2022)Berg, Hall, Bhalgat, Yang, Kirk, Shtedritski, and
  Bain}]{berg2022prompt}
Hugo Berg, Siobhan~Mackenzie Hall, Yash Bhalgat, Wonsuk Yang, Hannah~Rose Kirk,
  Aleksandar Shtedritski, and Max Bain. 2022.
\newblock \href {https://arxiv.org/abs/2203.11933} {A prompt array keeps the
  bias away: Debiasing vision-language models with adversarial learning}.
\newblock \emph{arXiv preprint arXiv:2203.11933}.

\bibitem[{Blodgett et~al.(2020)Blodgett, Barocas, Daum{\'e}~III, and
  Wallach}]{Blodgett2020}
Su~Lin Blodgett, Solon Barocas, Hal Daum{\'e}~III, and Hanna Wallach. 2020.
\newblock \href {https://doi.org/10.18653/v1/2020.acl-main.485} {Language
  (technology) is power: A critical survey of {``}bias{''} in {NLP}}.
\newblock In \emph{Proceedings of the 58th Annual Meeting of the Association
  for Computational Linguistics}, pages 5454--5476, Online. Association for
  Computational Linguistics.

\bibitem[{Bolukbasi et~al.(2016)Bolukbasi, Chang, Zou, Saligrama, and
  Kalai}]{Bolukbasi2016}
Tolga Bolukbasi, Kai-Wei Chang, James Zou, Venkatesh Saligrama, and Adam Kalai.
  2016.
\newblock \href {https://arxiv.org/abs/1607.06520} {Man is to computer
  programmer as woman is to homemaker? debiasing word embeddings}.
\newblock In \emph{Proceedings of the 30th International Conference on Neural
  Information Processing Systems}, NIPS'16, page 4356–4364, Red Hook, NY,
  USA. Curran Associates Inc.

\bibitem[{Bommasani et~al.(2021)Bommasani, Hudson, Adeli, Altman, Arora, von
  Arx, Bernstein, Bohg, Bosselut, Brunskill
  et~al.}]{bommasani2021opportunities}
Rishi Bommasani, Drew~A Hudson, Ehsan Adeli, Russ Altman, Simran Arora, Sydney
  von Arx, Michael~S Bernstein, Jeannette Bohg, Antoine Bosselut, Emma
  Brunskill, et~al. 2021.
\newblock \href {https://arxiv.org/abs/2108.07258} {On the opportunities and
  risks of foundation models}.
\newblock \emph{arXiv preprint arXiv:2108.07258}.

\bibitem[{Brown et~al.(2020)Brown, Mann, Ryder, Subbiah, Kaplan, Dhariwal,
  Neelakantan, Shyam, Sastry, Askell et~al.}]{brown2020language}
Tom Brown, Benjamin Mann, Nick Ryder, Melanie Subbiah, Jared~D Kaplan, Prafulla
  Dhariwal, Arvind Neelakantan, Pranav Shyam, Girish Sastry, Amanda Askell,
  et~al. 2020.
\newblock \href {https://arxiv.org/abs/2005.14165} {Language models are
  few-shot learners}.
\newblock \emph{Advances in Neural Information Processing Systems}, 33.

\bibitem[{{Cambridge University Dictionary}(2022)}]{_handsome}
{Cambridge University Dictionary}. 2022.
\newblock \href {https://dictionary.cambridge.org/dictionary/english/handsome}
  {handsome}.

\bibitem[{Chen et~al.(2021)Chen, Tworek, Jun, Yuan, de~Oliveira~Pinto, Kaplan,
  Edwards, Burda, Joseph, Brockman et~al.}]{Chen}
Mark Chen, Jerry Tworek, Heewoo Jun, Qiming Yuan, Henrique~Ponde
  de~Oliveira~Pinto, Jared Kaplan, Harrison Edwards, Yuri Burda, Nicholas
  Joseph, Greg Brockman, et~al. 2021.
\newblock \href {http://arxiv.org/abs/2107.03374} {Evaluating large language
  models trained on code}.
\newblock \emph{arXiv preprint arXiv:2107.03374}.

\bibitem[{Dev and Phillips(2019)}]{dev2019attenuating}
Sunipa Dev and Jeff Phillips. 2019.
\newblock \href {http://proceedings.mlr.press/v89/dev19a.html} {Attenuating
  bias in word vectors}.
\newblock In \emph{The 22nd International Conference on Artificial Intelligence
  and Statistics}, pages 879--887. PMLR.

\bibitem[{Devlin et~al.(2018)Devlin, Chang, Lee, and
  Toutanova}]{devlin2018bert}
Jacob Devlin, Ming-Wei Chang, Kenton Lee, and Kristina Toutanova. 2018.
\newblock \href {http://arxiv.org/abs/1810.04805} {Bert: Pre-training of deep
  bidirectional transformers for language understanding}.
\newblock \emph{arXiv preprint arXiv:1810.04805}.

\bibitem[{Fell and Dyban(2017)}]{fell2017against}
Elena~Vladimirovna Fell and Maria Dyban. 2017.
\newblock Against discrimination: equality act 2010 {(UK)}.
\newblock \emph{The European Proceedings of Social \& Behavioural Sciences
  (EpSBS). Vol. 19: Lifelong Wellbeing in the World (WELLSO 2016).—Nicosia,
  2017.}, 192016:188--194.

\bibitem[{Feyisetan et~al.(2020)Feyisetan, Ghanavati, and
  Thaine}]{Feyisetan2020}
Oluwaseyi Feyisetan, Sepideh Ghanavati, and Patricia Thaine. 2020.
\newblock \href {https://doi.org/10.1145/3336191.3371881} {\emph{Workshop on
  Privacy in NLP (PrivateNLP 2020)}}, pages 903--904. Association for Computing
  Machinery, New York, NY, USA.

\bibitem[{Fleiss(1971)}]{fleiss1971measuring}
Joseph~L Fleiss. 1971.
\newblock Measuring nominal scale agreement among many raters.
\newblock \emph{Psychological Bulletin}, 76(5):378.

\bibitem[{Gatt and Krahmer(2018)}]{Gatt2018}
Albert Gatt and Emiel Krahmer. 2018.
\newblock \href {https://dl.acm.org/doi/10.5555/3241691.3241693} {Survey of the
  state of the art in natural language generation: Core tasks, applications and
  evaluation}.
\newblock \emph{J. Artif. Int. Res.}, 61(1):65–170.

\bibitem[{Gaucher et~al.(2011)Gaucher, Friesen, and Kay}]{Gaucher2011}
Danielle Gaucher, Justin Friesen, and Aaron~C. Kay. 2011.
\newblock \href {https://doi.org/10.1037/a0022530} {{Evidence That Gendered
  Wording in Job Advertisements Exists and Sustains Gender Inequality}}.
\newblock \emph{Journal of Personality and Social Psychology}, 101(1):109--128.

\bibitem[{Gonen and Goldberg(2019)}]{Gonen2019}
Hila Gonen and Yoav Goldberg. 2019.
\newblock \href {https://doi.org/10.18653/v1/N19-1061} {Lipstick on a pig:
  {D}ebiasing methods cover up systematic gender biases in word embeddings but
  do not remove them}.
\newblock In \emph{Proceedings of the 2019 Conference of the North {A}merican
  Chapter of the Association for Computational Linguistics: Human Language
  Technologies, Volume 1 (Long and Short Papers)}, pages 609--614, Minneapolis,
  Minnesota. Association for Computational Linguistics.

\bibitem[{Joshi et~al.(2018)Joshi, Peters, and Hopkins}]{Joshi2018}
Vidur Joshi, Matthew Peters, and Mark Hopkins. 2018.
\newblock \href {https://doi.org/10.18653/v1/P18-1110} {Extending a parser to
  distant domains using a few dozen partially annotated examples}.
\newblock In \emph{Proceedings of the 56th Annual Meeting of the Association
  for Computational Linguistics (Volume 1: Long Papers)}, pages 1190--1199,
  Melbourne, Australia. Association for Computational Linguistics.

\bibitem[{Kirk et~al.(2021)Kirk, Jun, Iqbal, Benussi, Volpin, Dreyer,
  Shtedritski, and Asano}]{Kirk2021}
Hannah Kirk, Yennie Jun, Haider Iqbal, Elias Benussi, Filippo Volpin,
  Frederic~A. Dreyer, Aleksandar Shtedritski, and Yuki~M. Asano. 2021.
\newblock \href {http://arxiv.org/abs/2102.04130} {{Bias Out-of-the-Box: An
  Empirical Analysis of Intersectional Occupational Biases in Popular
  Generative Language Models}}.
\newblock \emph{Advances in Neural Information Processing Systems}, 34.

\bibitem[{Kurita et~al.(2019)Kurita, Vyas, Pareek, Black, and
  Tsvetkov}]{Kurita2019}
Keita Kurita, Nidhi Vyas, Ayush Pareek, Alan~W Black, and Yulia Tsvetkov. 2019.
\newblock \href {https://doi.org/10.18653/v1/W19-3823} {Measuring bias in
  contextualized word representations}.
\newblock In \emph{Proceedings of the First Workshop on Gender Bias in Natural
  Language Processing}, pages 166--172, Florence, Italy. Association for
  Computational Linguistics.

\bibitem[{Li et~al.(2020)Li, Geissinger, Ingram, and Fox}]{Li2020}
Liuqing Li, Jack Geissinger, William~A. Ingram, and Edward~A. Fox. 2020.
\newblock \href {https://doi.org/doi:10.2478/dim-2020-0003} {Teaching natural
  language processing through big data text summarization with problem-based
  learning}.
\newblock \emph{Data and Information Management}, 4(1):18--43.

\bibitem[{Liu et~al.(2020)Liu, Liu, Zhang, Chi, Shi, and
  Huang}]{liu-etal-2020-hiring}
Liting Liu, Jie Liu, Wenzheng Zhang, Ziming Chi, Wenxuan Shi, and Yalou Huang.
  2020.
\newblock \href {https://doi.org/10.18653/v1/2020.acl-main.281} {Hiring now: A
  skill-aware multi-attention model for job posting generation}.
\newblock In \emph{Proceedings of the 58th Annual Meeting of the Association
  for Computational Linguistics}, pages 3096--3104, Online. Association for
  Computational Linguistics.

\bibitem[{Liu et~al.(2021{\natexlab{a}})Liu, Yuan, Fu, Jiang, Hayashi, and
  Neubig}]{Liu2021}
Pengfei Liu, Weizhe Yuan, Jinlan Fu, Zhengbao Jiang, Hiroaki Hayashi, and
  Graham Neubig. 2021{\natexlab{a}}.
\newblock \href {http://arxiv.org/abs/2107.13586} {Pre-train, prompt, and
  predict: {A} systematic survey of prompting methods in natural language
  processing}.
\newblock \emph{arXiv preprint arXiv:2107.13586}.

\bibitem[{Liu et~al.(2021{\natexlab{b}})Liu, Jia, Wei, Xu, Wang, and
  Vosoughi}]{liu2021mitigating}
Ruibo Liu, Chenyan Jia, Jason Wei, Guangxuan Xu, Lili Wang, and Soroush
  Vosoughi. 2021{\natexlab{b}}.
\newblock \href {https://arxiv.org/abs/2104.14795} {Mitigating political bias
  in language models through reinforced calibration}.
\newblock In \emph{Proceedings of the AAAI Conference on Artificial
  Intelligence}, volume~35, pages 14857--14866.

\bibitem[{Lu et~al.(2020)Lu, Mardziel, Wu, Amancharla, and
  Datta}]{lu2020gender}
Kaiji Lu, Piotr Mardziel, Fangjing Wu, Preetam Amancharla, and Anupam Datta.
  2020.
\newblock \href {http://arxiv.org/abs/1807.11714} {Gender bias in neural
  natural language processing}.
\newblock In \emph{Logic, Language, and Security}, pages 189--202. Springer.

\bibitem[{Margoni(2019)}]{Margoni2019}
Thomas Margoni. 2019.
\newblock \href {https://doi.org/10.2139/ssrn.3299523} {{Artificial
  Intelligence, Machine Learning and EU Copyright Law: Who Owns AI?}}
\newblock \emph{SSRN Electronic Journal}, 12(December).

\bibitem[{Mohammad(2018)}]{mohammad2018obtaining}
Saif Mohammad. 2018.
\newblock \href {https://aclanthology.org/P18-1017} {Obtaining reliable human
  ratings of valence, arousal, and dominance for 20,000 english words}.
\newblock In \emph{Proceedings of the 56th Annual Meeting of the Association
  for Computational Linguistics (Volume 1: Long Papers)}, pages 174--184.

\bibitem[{{ONS}(2018)}]{ONS2018}
{ONS}. 2018.
\newblock \href
  {https://www.ons.gov.uk/employmentandlabourmarket/peopleinwork/employmentandemployeetypes/adhocs/008439employmntbydetailedoccupationandindustrybysexandageforgreatbritainukandconstituentcountries}
  {{UK} office for national statistics:employment by detailed occupation and
  industry by sex and age for great britain, {UK} and constituent countries}.
\newblock Accessed: 2022-02-25.

\bibitem[{Ouyang et~al.(2022)Ouyang, Wu, Jiang, Almeida, Wainwright, Mishkin,
  Zhang, Agarwal, Slama, Ray et~al.}]{ouyang2022training}
Long Ouyang, Jeff Wu, Xu~Jiang, Diogo Almeida, Carroll~L Wainwright, Pamela
  Mishkin, Chong Zhang, Sandhini Agarwal, Katarina Slama, Alex Ray, et~al.
  2022.
\newblock \href {https://arxiv.org/abs/2203.02155} {Training language models to
  follow instructions with human feedback}.
\newblock \emph{arXiv preprint arXiv:2203.02155}.

\bibitem[{Sap et~al.(2017)Sap, Prasettio, Holtzman, Rashkin, and
  Choi}]{Sap2017}
Maarten Sap, Marcella~Cindy Prasettio, Ari Holtzman, Hannah Rashkin, and Yejin
  Choi. 2017.
\newblock \href {https://doi.org/10.18653/v1/D17-1247} {Connotation frames of
  power and agency in modern films}.
\newblock In \emph{Proceedings of the 2017 Conference on Empirical Methods in
  Natural Language Processing}, pages 2329--2334, Copenhagen, Denmark.
  Association for Computational Linguistics.

\bibitem[{Schmader et~al.(2007)Schmader, Whitehead, and Wysocki}]{Schmader2007}
Toni Schmader, Jessica Whitehead, and Vicki~H Wysocki. 2007.
\newblock \href {https://doi.org/10.1007/s11199-007-9291-4} {{A Linguistic
  Comparison of Letters of Recommendation for Male and Female Chemistry and
  Biochemistry Job Applicants}}.
\newblock \emph{Sex Roles}, pages 509--514.

\bibitem[{Sen et~al.(2021)Sen, Samory, Floeck, Wagner, and
  Augenstein}]{sen2021does}
Indira Sen, Mattia Samory, Fabian Floeck, Claudia Wagner, and Isabelle
  Augenstein. 2021.
\newblock How does counterfactually augmented data impact models for social
  computing constructs?
\newblock \emph{arXiv preprint arXiv:2109.07022}.

\bibitem[{Sheng et~al.(2019)Sheng, Chang, Natarajan, and Peng}]{Sheng2019}
Emily Sheng, Kai-Wei Chang, Premkumar Natarajan, and Nanyun Peng. 2019.
\newblock \href {https://doi.org/10.18653/v1/D19-1339} {The woman worked as a
  babysitter: On biases in language generation}.
\newblock In \emph{Proceedings of the 2019 Conference on Empirical Methods in
  Natural Language Processing and the 9th International Joint Conference on
  Natural Language Processing (EMNLP-IJCNLP)}, pages 3407--3412, Hong Kong,
  China. Association for Computational Linguistics.

\bibitem[{Smith and Williams(2021)}]{smith2021hi}
Eric~Michael Smith and Adina Williams. 2021.
\newblock \href {http://arxiv.org/abs/2109.03300} {Hi, my name is
  \uppercase{M}artha: Using names to measure and mitigate bias in generative
  dialogue models}.
\newblock \emph{arXiv preprint arXiv:2109.03300}.

\bibitem[{Smith et~al.(2022)Smith, Patwary, Norick, LeGresley, Rajbhandari,
  Casper, Liu, Prabhumoye, Zerveas, Korthikanti et~al.}]{Smith2022}
Shaden Smith, Mostofa Patwary, Brandon Norick, Patrick LeGresley, Samyam
  Rajbhandari, Jared Casper, Zhun Liu, Shrimai Prabhumoye, George Zerveas,
  Vijay Korthikanti, et~al. 2022.
\newblock \href {http://arxiv.org/abs/2201.11990} {Using deepspeed and megatron
  to train megatron-turing {NLG} 530b, {A} large-scale generative language
  model}.
\newblock \emph{arXiv preprint arXiv:2201.11990}.

\bibitem[{Solaiman et~al.(2019)Solaiman, Brundage, Clark, Askell, Herbert-Voss,
  Wu, Radford, Krueger, Kim, Kreps et~al.}]{solaiman2019release}
Irene Solaiman, Miles Brundage, Jack Clark, Amanda Askell, Ariel Herbert-Voss,
  Jeff Wu, Alec Radford, Gretchen Krueger, Jong~Wook Kim, Sarah Kreps, et~al.
  2019.
\newblock \href {http://arxiv.org/abs/1908.09203} {Release strategies and the
  social impacts of language models}.
\newblock \emph{arXiv preprint arXiv:1908.09203}.

\bibitem[{Solaiman and Dennison(2021)}]{solaiman2021process}
Irene Solaiman and Christy Dennison. 2021.
\newblock \href
  {https://proceedings.neurips.cc/paper/2021/hash/2e855f9489df0712b4bd8ea9e2848c5a-Abstract.html}
  {Process for adapting language models to society (palms) with values-targeted
  datasets}.
\newblock \emph{Advances in Neural Information Processing Systems}, 34.

\bibitem[{Somers et~al.(1997)Somers, Black, Nivre, Lager, Multari, Gilardoni,
  Ellman, and Rogers}]{10.3115/974557.974597}
Harold Somers, Bill Black, Joakim Nivre, Torbj\"{o}rn Lager, Annarosa Multari,
  Luca Gilardoni, Jeremy Ellman, and Alex Rogers. 1997.
\newblock \href {https://doi.org/10.3115/974557.974597} {Multilingual
  generation and summarization of job adverts: The tree project}.
\newblock In \emph{Proceedings of the Fifth Conference on Applied Natural
  Language Processing}, ANLC '97, page 269–276, USA. Association for
  Computational Linguistics.

\bibitem[{Stanczak and Augenstein(2021)}]{Stanczak2021}
Karolina Stanczak and Isabelle Augenstein. 2021.
\newblock \href {http://arxiv.org/abs/2112.14168} {A survey on gender bias in
  natural language processing}.
\newblock \emph{arXiv preprint arXiv:2112.14168}.

\bibitem[{Sun et~al.(2019)Sun, Gaut, Tang, Huang, ElSherief, Zhao, Mirza,
  Belding, Chang, and Wang}]{sun2019mitigating}
Tony Sun, Andrew Gaut, Shirlyn Tang, Yuxin Huang, Mai ElSherief, Jieyu Zhao,
  Diba Mirza, Elizabeth Belding, Kai-Wei Chang, and William~Yang Wang. 2019.
\newblock \href {http://arxiv.org/abs/1906.08976} {Mitigating gender bias in
  natural language processing: Literature review}.
\newblock \emph{arXiv preprint arXiv:1906.08976}.

\bibitem[{Veale(2016)}]{Veale2016}
Tony Veale. 2016.
\newblock \href {https://doi.org/10.1350/jcla.2009.73.2.552} {{Round Up The
  Usual Suspects: Knowledge-Based Metaphor Generation}}.
\newblock \emph{Proceedings of The Fourth Workshop on Metaphor in NLP, San
  Diego, USA}, pages 34--41.

\bibitem[{Weidinger et~al.(2021)Weidinger, Mellor, Rauh, Griffin, Uesato,
  Huang, Cheng, Glaese, Balle, Kasirzadeh et~al.}]{Weidinger2021}
Laura Weidinger, John Mellor, Maribeth Rauh, Conor Griffin, Jonathan Uesato,
  Po{-}Sen Huang, Myra Cheng, Mia Glaese, Borja Balle, Atoosa Kasirzadeh,
  et~al. 2021.
\newblock \href {http://arxiv.org/abs/2112.04359} {Ethical and social risks of
  harm from language models}.
\newblock \emph{arXiv preprint arXiv:2112.04359}.

\bibitem[{Zhang et~al.(2018)Zhang, Lemoine, and Mitchell}]{zhang2018mitigating}
Brian~Hu Zhang, Blake Lemoine, and Margaret Mitchell. 2018.
\newblock \href {https://dl.acm.org/doi/abs/10.1145/3278721.3278779}
  {Mitigating unwanted biases with adversarial learning}.
\newblock In \emph{Proceedings of the 2018 AAAI/ACM Conference on AI, Ethics,
  and Society}, pages 335--340.

\end{thebibliography}

\appendix
\clearpage
\section{Ethical Considerations and Risks}
\label{appendixA:ethics}

 \paragraph{Misuse} Our paper highlights the risk of generative language models outputting biased text which propagates or amplifies societal biases. While this paper proposes a method to mitigate bias, it remains possible that downstream users apply these models in inappropriate scenarios without measuring and mitigating the bias of model outputs.

\paragraph{Viability} It is possible that fine-tuning will not be viable in all domains. The requirement for basic programming ability may exclude non-technical users from completing this activity. Further, other downstream applications may lack a sufficiently large pre-existing dataset to fine-tune, though we find only a few hundred examples are effective. 

\paragraph{GPT-3 Licence Terms} Our application fits within the described intended use of GPT-3 as a ``narrow generative use case''. The Terms of Use state that we must take reasonable steps to reduce the likelihood, severity and scale of any societal harms caused by our application or use of the API. Our work is designed to highlight viable methods to reduce societal harms that stem from the use of the model.

\paragraph{Cost} The total computational cost of running our experiments was \$362.84. Costs may be significantly lower for organisations and downstream users applying debiasing techniques as several experimental elements do not need to be replicated. 

\section{Further Detail on GPT-3 Hyperparameters}
\label{sec:gpt_hyperparams}
\begin{figure}[t]
    \centering
    \includegraphics[width = \columnwidth]{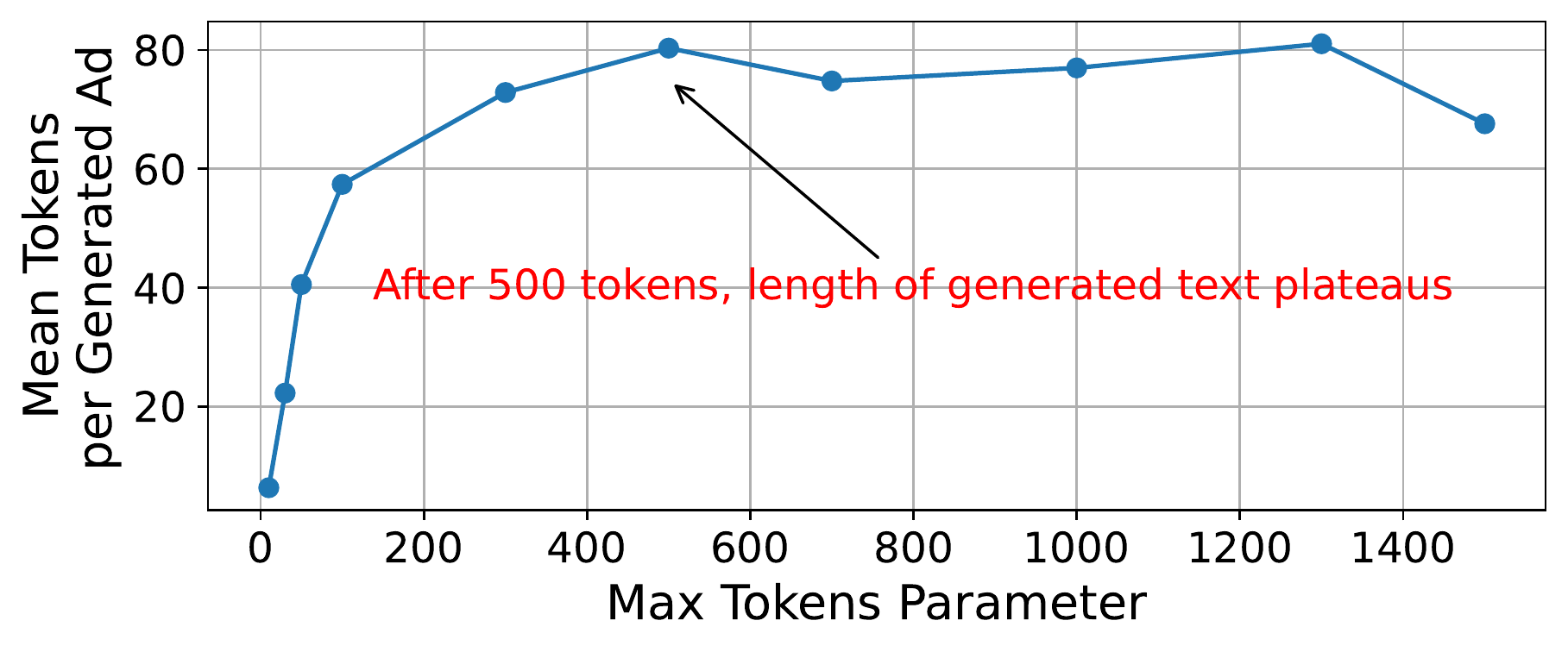}
    \caption{\textbf{Determining ad length.} Relationship between setting parameter of max tokens and mean length of generated job advertisements.}
    \label{fig:token_plot}
\end{figure}

For all experiments we used the Davinci GPT-3 model from OpenAI with the following parameters: 
\begin{itemize}
\setlength\itemsep{0.2em}
    \item  \texttt{max\_tokens} = 500
    \item \texttt{temperature} = 1
    \item \texttt{top\_p} = 1
    \item \texttt{n} = 1
    \item \texttt{stop} = null
    \item \texttt{presence\_penalty} = 0
    \item \texttt{best\_of} = 1
\end{itemize}
The value of 500 max tokens was determined experimentally by progressively allowing the model to use more tokens per completion with the following zero-shot prompt: ``Write a job advertisement for a \{job\}.'' and observing how that affects the number of words generated (see \cref{fig:token_plot}).

\section{Job Selection}
\label{sec:job_selection_detail}
To select candidate jobs for experiments, we use the list of jobs in the UK ASHE report \cite{ONS2018}. We filter jobs to $\ge 40{,}000$ employees nationwide (prevalence) and at least $1{,}000$ ads posted on a popular UK job site (relevance) to focus on jobs and ads which have an impact on a large number of people.
We translate job titles to accessible versions (e.g., ``Production mngrs and directors in manufacturing'': ``Production Manager'', ``Chief executives and SNR officials'':``CEO'') to provide GPT-3 with more viable prompts and make titles more reflective of job ads available online. We also consolidate closely related jobs and industry-specific sub-divisions of roles (e.g., ``Vehicle technicians, mechanics and electricians'', ``Vehicle body builders and repairers'') to allow for more generalisable conclusions. Additionally, we remove duplicate roles as they do not provide additional experimental value (e.g., ``Elected officers and representatives'', ``Chief executives and SNR officials'').
To estimate GPT-3's priors of bias between these remaining jobs and gender, we run tests with several proposed prompts:
\begin{enumerate}
    \item ``The \{job\} is a man/woman. The \{job\} is a [token]''
    \item ``The \{job\} is a woman/man. The \{job\} is a [token]''
    \item ``The \{job\} is a [token]''
    \item ``What gender is the \{job\}? The \{job\} is typically a [token]''
    \item ``What gender is the \{job\}? The \{job\} is a [token]''
\end{enumerate}
Out of these, we select prompt 5 which provided the highest proportion of gendered tokens. Each completion is repeated $1{,}000$ times, where completions are limited to 1 token to context-force the most likely next token. Based on these completions, we calculate two metrics:

\paragraph{Genderedness, $G$} The proportion of returned tokens which are gendered ($T \in$ {GENDERED}=[``male'', ``man'', ``masculine'', ``female'', ``women'', ``women'', \ldots]) out of $1{,}000$ completions:
\begin{equation}
    G = \frac{\sum_{T\in C}T[T_i \in \mathrm{GENDERED}]}{\sum_{T\in C} T_i}
\end{equation}

\paragraph{Bias Margin, $B$} The overrepresentation factor of female tokens in all gendered tokens relative to the equal distribution (i.e., 50:50 representation across gendered tokens):
\begin{equation}
    B = \frac{G*0.5 - \sum_{T\in C}T_i[T_i = \mathrm{FEMALE}]}{G * 0.5}
\end{equation}
Where $B\in[-1,0]$ if the job is female-biased and $B\in[0,1]$ if male-biased.

The selected jobs by prevalence, relevance and bias margin are shown in \cref{tab:job_selection}.

\begin{table}
\caption{\textbf{ONS UK labour market statistics}. Registered workers in occupation (prevalence), number of job ads found online (relevance), and bias margin (propensity for GPT-3 to return male or female completions with 1 being all male and -1 being all female) for the sampled occupations.}
\label{tab:job_selection}
\footnotesize
\centering
\resizebox{\columnwidth}{!}{
\begin{tabular}{lccc} 
\toprule
\textbf{Job}           & \textbf{Prevalence} & \textbf{Relevance} & \textbf{Bias}  \\ 
\hline
\multicolumn{4}{c}{{\cellcolor[rgb]{0.753,0.753,0.753}}Female-Biased}              \\ 
\hline
Nurse                  & 622,998             & 43,259             & -1.00          \\
Housekeeper            & 41,626              & 3,088              & -1.00          \\
Occupational Therapist & 43,888              & 2,990              & -1.00          \\
Secretary              & 195,375             & 2,235              & -0.99          \\
Social Worker          & 104,992             & 4,721              & -0.99          \\ 
\hline
\multicolumn{4}{c}{{\cellcolor[rgb]{0.753,0.753,0.753}}Male-Biased}                \\ 
\hline
Plumber                & 184,707             & 1,598              & 1.00           \\
Engineer               & 133,662             & 57,958             & 0.92           \\
Carpenter              & 233,387             & 1,444              & 1.00           \\
Electrician            & 241,738             & 3,045              & 1.00           \\
Software Developer     & 303,330             & 2,306              & 0.98           \\ 
\hline
\multicolumn{4}{c}{{\cellcolor[rgb]{0.753,0.753,0.753}}Neutral}                    \\ 
\hline
Artist                 & 50,744              & 1,286              & 0.02           \\
Tester                 & 78,221              & 2,277              & -0.03          \\
Administrator          & 814,583             & 22,017             & 0.07           \\
Project Manager        & 72,785              & 9,565              & 0.08           \\
Writer                 & 86,145              & 1,359              & 0.13           \\
\bottomrule
\end{tabular}
}
\end{table}

\section{Neutral and Engineered Prompts} 
\label{sec:prompt_detail}
GPT-3 displays strong zero-shot abilities \cite{brown2020language}, i.e., using a simple instruction or ``prompt'' as input, the model will extend or complete the text accordingly without any pre-defined examples. Prompt-engineering thus refers to manipulations and perturbations of this prompt to context-force the desired output behaviour \cite{Liu2021}. In contrast to zero-shot, GPT-3 can be fine-tuned over a dataset with desired input-output pairs \cite{brown2020language}. 
To conduct the experiment to compare neutral and diversity-encouraging prompts, we compile a list of 18 prompts. Nine of them are designated ``neutral'' and used as our ``zero-shot'' prompts. These simply specify a task of generating an ad for a given job but are syntactically varied. The other nine prompts are ``equality and diversity prompts'', which we call ``engineered'' prompts. \cref{tab:prompt_table} displays all 18 prompts with their respective bias and realism scores.

\begin{table}
\centering
\caption{\textbf{Neutral and engineered prompts}. Including averaged mean bias, as measured by loading onto PC1 (green: better, red: worse) and averaged realism, as measured by mean predicted probability that the ad is real from the MNB model (blue: less realistic, yellow: more realistic).}
\label{tab:prompt_table}
\footnotesize
\setlength{\extrarowheight}{0pt}
\addtolength{\extrarowheight}{\aboverulesep}
\addtolength{\extrarowheight}{\belowrulesep}
\setlength{\aboverulesep}{0pt}
\setlength{\belowrulesep}{0pt}
\begin{tabular}{p{4cm}p{1cm}p{1cm}} 
\toprule
\textbf{Prompt Template}                                                           & \textbf{Bias}                            & \textbf{Realism}                      \\ 
\hline
\rowcolor[rgb]{0.753,0.753,0.753} Neutral Prompts (Mean)                           & 0.059                                     & 0.004                                      \\ 
\hline
{[}"Compose a job ad for a \{job\}."]                                              & {\cellcolor[rgb]{0.961,0.78,0.765}}0.061  & {\cellcolor[rgb]{0.788,0.859,0.973}}0.004  \\
{[}"Write a job ad for a \{job\}."]                                                & {\cellcolor[rgb]{0.969,0.816,0.804}}0.061 & 0.006                                      \\
{[}"Write a job advertisement for a \{job\}."]                                     & {\cellcolor[rgb]{0.976,0.863,0.855}}0.060 & {\cellcolor[rgb]{0.686,0.788,0.961}}0.003  \\
{[}"Compose a job advertisement for a \{job\}."]                                   & {\cellcolor[rgb]{0.988,0.929,0.922}}0.060 & {\cellcolor[rgb]{0.984,0.988,0.996}}0.006  \\
{[}"Generate a job ad for a \{job\}."]                                             & {\cellcolor[rgb]{0.992,0.996,0.996}}0.059 & {\cellcolor[rgb]{0.694,0.796,0.961}}0.003  \\
{[}"Generate a job advertisement for a \{job\}."]                                  & {\cellcolor[rgb]{0.929,0.969,0.949}}0.058 & 0.006                                      \\
{[}"Write a job advertisement for the following profession: \{job\}."]             & {\cellcolor[rgb]{0.918,0.965,0.941}}0.058 & {\cellcolor[rgb]{0.706,0.804,0.961}}0.003  \\
{[}"Compose a job advertisement for the following profession: \{job\}."]           & {\cellcolor[rgb]{0.91,0.965,0.937}}0.058  & {\cellcolor[rgb]{0.663,0.776,0.957}}0.003  \\
{[}"Generate a job advertisement for the following profession: \{job\}."]          & {\cellcolor[rgb]{0.878,0.949,0.914}}0.057 & {\cellcolor[rgb]{0.643,0.761,0.957}}0.002  \\ 
\hline
\rowcolor[rgb]{0.753,0.753,0.753} Engineered Prompts (Mean)                        & 0.058                                     & 0.027                                      \\ 
\hline
{[}"Write a job ad without ay gender bias for a \{job\}."]                         & {\cellcolor[rgb]{0.902,0.486,0.451}}0.063 & {\cellcolor[rgb]{1,1,0.996}}0.007          \\
{[}"We are fair ad equal opportunities employer. Write a job ad for a \{job\}."]   & {\cellcolor[rgb]{0.925,0.592,0.565}}0.062 & {\cellcolor[rgb]{1,1,0.992}}0.008          \\
{[}"Write a gender neutral job ad for a \{job\}."]                                 & {\cellcolor[rgb]{0.937,0.659,0.635}}0.062 & {\cellcolor[rgb]{1,0.996,0.976}}0.012      \\
{[}"Compose a unbiased job ad for a \{job\}."]                                     & {\cellcolor[rgb]{0.941,0.678,0.655}}0.062 & {\cellcolor[rgb]{0.824,0.882,0.976}}0.004  \\
{[}"Write a unbiased job ad for a \{job\}."]                                       & {\cellcolor[rgb]{1,0.988,0.984}}0.060     & {\cellcolor[rgb]{0.816,0.875,0.976}}0.004  \\
{[}"Write a job ad for a \{job\} which appeals equally to men ad women."]          & {\cellcolor[rgb]{0.961,0.984,0.973}}0.059 & {\cellcolor[rgb]{1,1,0.988}}0.009          \\
{[}"We are committed to diversity in our firm, write a job ad for a new \{job\}."] & {\cellcolor[rgb]{0.894,0.957,0.925}}0.057 & {\cellcolor[rgb]{1,0.918,0.678}}0.075      \\
{[}"Write a job ad for a \{job\} for a firm focused on diversity in hiring."]      & {\cellcolor[rgb]{0.745,0.894,0.82}}0.054  & {\cellcolor[rgb]{1,0.898,0.6}}0.090        \\
{[}"We are focused on hiring minority groups, write a job ad for a \{job\}."]      & {\cellcolor[rgb]{0.341,0.733,0.541}}0.046 & {\cellcolor[rgb]{1,0.965,0.859}}0.036      \\
\bottomrule
\end{tabular}
\end{table}

\section{Constructing Bias Measures}
\label{sec:bias_detail}
We provide a detailed summary of the individual bias measures used in our composite bias score. Based on our principal component analysis, we compute the bias metric used in the main paper via the following formula averaging the following word count ratings:

$$
\frac{\sum (\mathrm{NRC}, \mathrm{Male}, \mathrm{Genderedness}, \mathrm{Superlative})}{4 \cdot N_{words}}
$$

\paragraph{Gender-Coded Word Prevalence}
This method \cite{Gaucher2011} is operationalised through a set of masculine- and feminine-themed words in the context of job ads. ``Adventurous'' and ``stubborn'' are coded as masculine words while ``affectionate'' and ``kind'' are coded as feminine words. This research provides us with 42 masculine and 40 feminine words, with a wider set of potential words permeating from these (i.e. ``Compet*'' which may manifest itself as competitive, competition and so on). Our measure counts the prevalence of these words in a given text. The calculation is:

$$
\frac{n_\mathrm{biased\, words}}{n_\mathrm{words}}
$$

\paragraph{Superlative Prevalence}
This measure is based on a correlation identified between ``standout'' words to describe a job candidate and research skill when describing that candidate \cite{Schmader2007}. A particular distinction is made between positive (standout) superlatives and negative (grindstone) superlatives and their differential use to describe men and women. In our experiment, we measure the prevalence of a set of superlatives provided by \citet{Veale2016}. The calculation is:

$$\frac{n_\mathrm{superlatives}}{n_\mathrm{words}}$$

\paragraph{Gender-Laden Scoring}
A previous study provides a list of $2{,}311$ words, based on an analysis of 32 properties related to a set of norms \cite{Sap2017}. In this study, words are scored for their ``gender-ladenness'' and ``gender replication''. Our study takes a count of the former, measuring their unweighted prevalence to make it comparable to the other bias measures. The calculation is:

$$\frac{n_\mathrm{biased \, words}}{n_\mathrm{words}}$$

\paragraph{Connotation Frames}
This measure is based on the concept of power and agency connotation frames \cite{Sap2017}. Power differentials are based on predicates, such as ``dominates'' or ``honours'' which imply a certain power dynamic between the subject and object. Agency is attributed to the agent of the verb. A set of transitive verbs ($1{,}700$ for power differentials and $2{,}000$ for agency) have been annotated in a previous study on modern films and operationalised in our scoring \cite{Sap2017}. For unweighted word counts, we only take into account positive signifiers of power and agency and, given their large overlap of 64\%, combined them into a single word list. The calculation is:

$$\frac{n_\mathrm{positive \, signifiers}}{n_\mathrm{words}}$$

\paragraph{NCR VAD Lexicon}
This measure is based on a list of words rated on the emotional dimensions of valence, arousal, and dominance which has been used in gender bias research. In particular, weakness (low dominance), passiveness (low arousal or agency), and badness (valence) may be associated with a female stereotype \cite{Stanczak2021}. Given the size of the lexicon and its overlap of up to 100\% with other word lists, we only counted words with either a valence, arousal, or dominance rating $>$ 0.75 on a scale from 0 to 1. The calculation is:

$$\frac{n_{\mathrm{words \, with\, VAD} > 0.75}}{n_\mathrm{words}}$$

\section{Abnormalities in Generated Ads}
\label{sec:bizarre_detail}
We present some details of unrealistic job ads generated in our experiments. These examples point to the need for human-directed assessments of generated completions and suggest GPT-3's shortcomings in some aspects of both zero-shot, prompt-engineered and fine-tuned completions.

Some zero-shot completions, which we generated for our baseline condition, seem to state what was expected of a job ad, rather than being an ad itself:
\begin{itemize}
    \item \texttt{prompt} = ``Write a job ad for a social worker.''
    \item \texttt{completion} = ``The job ad should list the responsibilities and qualifications of a social worker.''
\end{itemize}
Other zero-shot, baseline completions describe the job itself, rather than being an advertisement:
\begin{itemize}
    \item \texttt{prompt} = ``Write a job ad for an artist.''
    \item \texttt{completion} = ``An artist is someone who creates artwork. They could be a painter, sculptor, or musician. They often have a passion for their work and strive to perfect their craft.''
\end{itemize}
Further, with some ads generated under the prompt-engineering condition, the model simply echoes the prompt:
\begin{itemize}
    \item \texttt{prompt} = ``Write a gender neutral job ad for an engineer.''
    \item \texttt{completion} = ``We are looking for an engineer to work on our team. This is a gender neutral job ad.''
\end{itemize}
In the same prompt-engineering experiment, we also encounter direct and overt references to diversity in hiring:
\begin{itemize}
    \item \texttt{prompt} = ``We are focused on hiring minority groups, write a job ad for an artist.''
    \item \texttt{completion} = ``We are focused on hiring minorities for our upcoming artist position. If you are an artist of color, please apply!''
\end{itemize}
In the fine-tuning condition, some ads do not match the specified job description:
\begin{itemize}
    \item \texttt{prompt} = ``Write a job ad for a plumber.''
    \item \texttt{completion} = ``(...) Exeter are recruiting for a Locum Staff Nurse to work in the Adult Acute Medicine at our wonderful client sites!(...)''
\end{itemize} Moreover, some fine-tuned ads have the right structure and described the correct job, but mention features of the perfect candidate completely unrelated to the nature of the job, which would not be mentioned in a real ad:
\begin{itemize}
    \item \texttt{prompt} = ``Write a job ad for an engineer.''
    \item \texttt{completion} = ``Job Title: Team Support Engineer(...) About You:(...) Willing to work as a Christian without compromising your values.''\end{itemize}
Finally, some fine-tuned completions contain repeated strings, either consisting of just a few letters, or whole words:
\begin{itemize}
    \item \texttt{prompt} = ``Write a job ad for a writer.''
    \item \texttt{completion 1} = ``Writer required \\
    Maternity Cover \\ Closing Date: 26/01/2022 \\ Job Type: Contract\\Salary: £30.00 per hour\\Contract length: 3 months\\COVID-1901VIDVIDVIDVIDVIDVIDVIDVID\\VIDVIDVIDVIDVIDVIDVIDVIDVIDVIDVID\\VIDVIDVIDVIDVIDVIDVIDVIDVIDVIDVID\\VIDVIDVIDVIDVIDVIDVIDVIDVIDVID(...)''
    \item \texttt{completion 2} = ``Looking for enthusiastic, hardworking writers and bloggers to join our team(...). Apply now and get the bonus bonus bonus bonus bonus bonus bonus bonus bonus bonus bonus bonus bonus bonus bonus bonus bonus bonus bonus bonus bonus bonus bonus bonus bonus bonus bonus bonus bonus bonus bonus bonus bonus bonus bonus bonus bonus bonus bonus bonus bonus bonus (...)''
\end{itemize}

\end{document}